\documentclass[conference]{IEEEtran}
\IEEEoverridecommandlockouts
\usepackage{cite}
\usepackage{hyperref}
\usepackage{amsmath,amssymb,amsfonts}
\usepackage{algorithmic}
\usepackage{graphicx}
\usepackage{textcomp}
\usepackage{xcolor}
\def\BibTeX{{\rm B\kern-.05em{\sc i\kern-.025em b}\kern-.08em
    T\kern-.1667em\lower.7ex\hbox{E}\kern-.125emX}}
\begin{document}

\title{MROSS: Multi-Round Region-based Optimization for Scene Sketching\\

\thanks{* Corresponding author.}
}

\author{\IEEEauthorblockN{ Yiqi Liang, Liu Ying*, Dandan Long, Ruihui Li }
\IEEEauthorblockA{\textit{College of Computer Science and Electronic Engineering} \\
\textit {Hunan University}\\
Changsha, China \\
yiqiliang@hnu.edu.cn, liu$\_$ying@hnu.edu.cn, ddlong@hnu.edu.cn, liruihui@hnu.edu.cn}
}

\maketitle

\begin{abstract}
Scene sketching is to convert a scene into a simplified, abstract representation that captures the essential elements and composition of the original scene. It requires a semantic understanding of the scene and consideration of different regions within the scene. Since scenes often contain diverse visual information across various regions, such as foreground objects, background elements, and spatial divisions, dealing with these different regions poses unique difficulties. In this paper, we define a sketch as some sets of Bézier curves because of their smooth and versatile characteristics. We optimize different regions of input scene in multiple rounds. In each optimization round, the strokes sampled from the next region can seamlessly be integrated into the sketch generated in the previous optimization round. We propose an additional stroke initialization method to ensure the integrity of the scene and the convergence of optimization. A novel CLIP-based Semantic Loss and a VGG-based Feature Loss are utilized to guide our multi-round optimization. Extensive experimental results on the quality and quantity of the generated sketches confirm the effectiveness of our method.
\end{abstract}

\begin{IEEEkeywords}
sketch, vector image, optimization, image processing
\end{IEEEkeywords}

\section{Introduction}
\label{sec:intro}
Scene sketching refers to the process of creating rough sketches or drawings to visually represent a scene or environment. It is commonly used in various fields, including art, design, architecture, film, and animation. 

Scene sketching offers advantage of conveying information quickly and efficiently. It provides a concise visual summary that allows viewers to grasp the overall layout, spatial relationships, and key features at a glance. Additionally, scene sketching facilitates the creative process by enabling artists and designers to visualize their ideas and concepts~\cite{wu2023sketchscene, koley2023picture, liu2024sketchdream}.

However, generating a scene sketch is highly challenging, as it requires the ability to understand and depict the visual characteristics of the scene with complex subjects and interactions.~\cite{azadi2018multi, chen2018cartoongan, isola2017image, liu2024sketchdream} often rely on explicit sketch datasets for training. The sketches of them are often simplified and abstract expressions of the original images, with a fixed style or preset. It is difficult to balance visual effect of sketches, producing visual appeal and aesthetics.

\begin{figure}[t]
        \centering
        \includegraphics[width=\columnwidth]{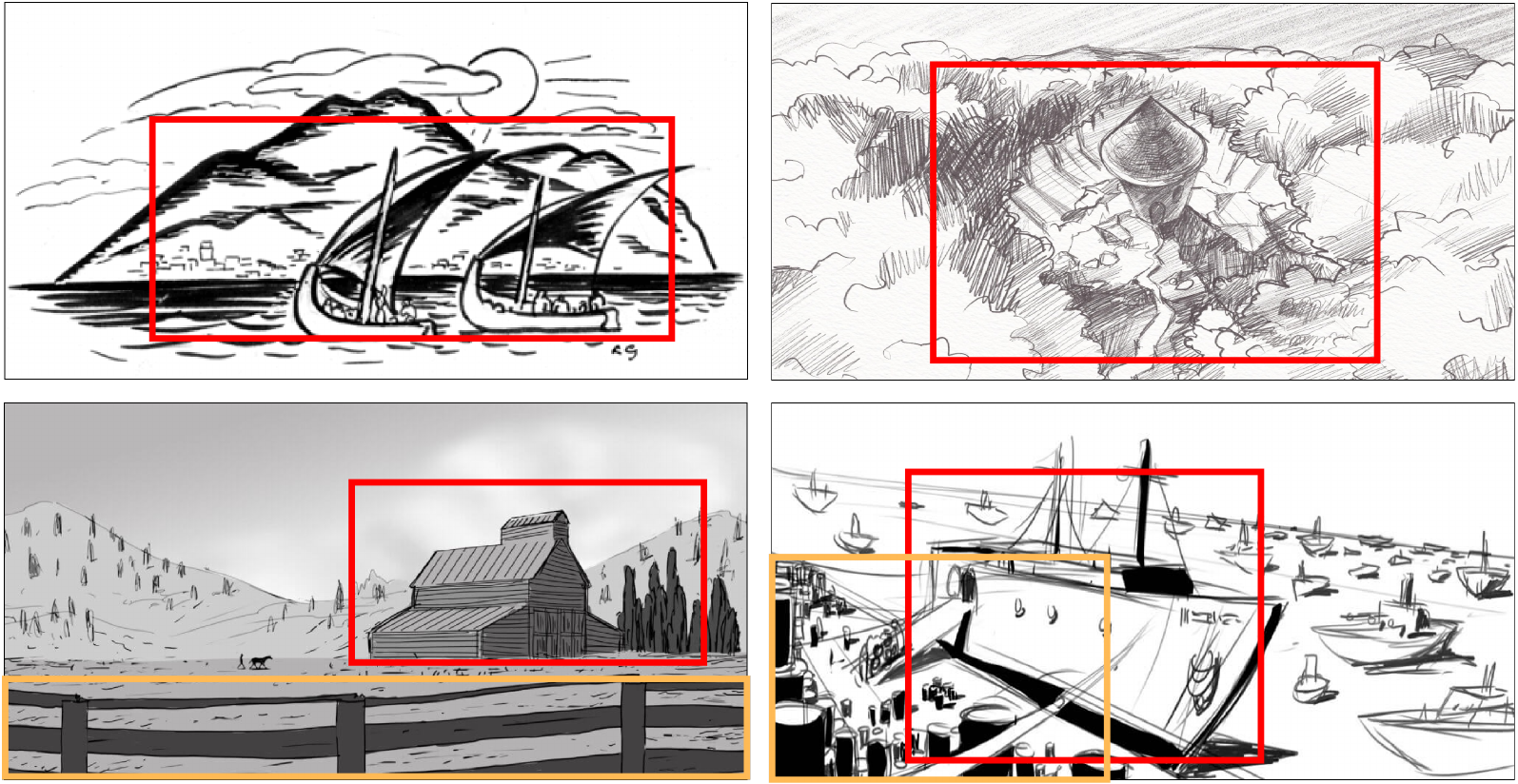} 
        \caption{Drawings of different scenes by different artists. Notice the significant differences in level of abstraction between different regions of the drawings.}
        \label{arts}
\vspace{-0.4cm}
    \end{figure}

Besides, different regions within a scene may have varying levels of importance or prominence as seen in Fig.~\ref{arts}. For example, foreground objects or focal points might require more attention to detail and precision in sketching, while background elements may be more loosely represented. Some works can achieve this with flexibility. However, these works focus specifically on the task of object sketching~\cite{liao2024freehand, vinker2022clipasso} or portrait sketching~\cite{berger2013style, guo2024face}, and often simply use the number of strokes to define the effect of their sketches.

To address the above two issues, we introduce a scene sketching method based on regions with multi-round optimization. We utilize the black parametric Bézier curves as our fundamental shape primitive for strokes of a sketch and optimized them by a pre-trained CLIP-ViT model~\cite{radford2021learning, dosovitskiy2020image} and VGG16 model~\cite{simonyan2014very}. 

\begin{figure*}[t]
    \centering
    \includegraphics[width=\textwidth]{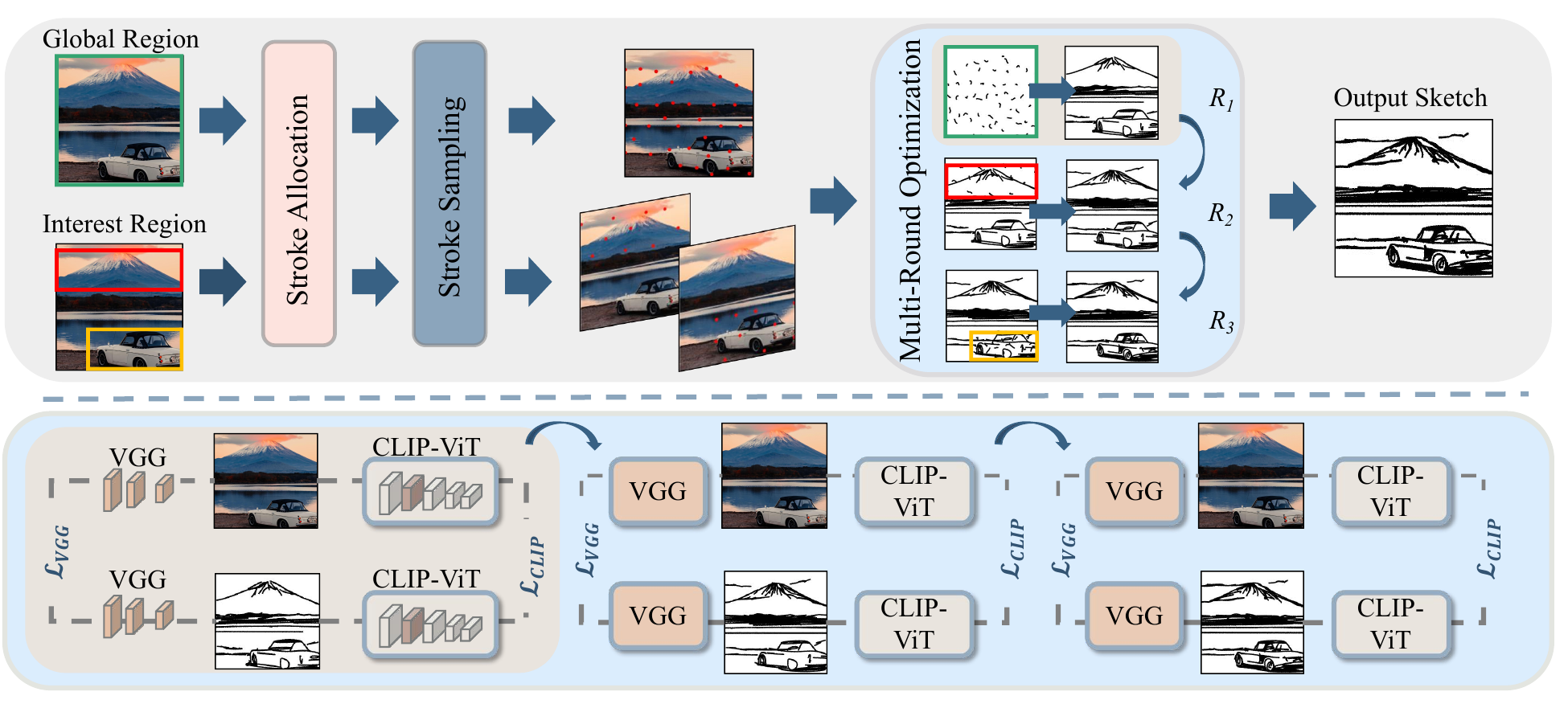} 
        \caption{Method overview – Given a scene photograph, and the selected regions, stroke initialization (stroke allocation and stroke sampling) is used to determine the initial strokes locations (red points) of different regions. These initial stroke locations will be converted into Bézier curves (black curves) as input of our multi-round optimization. In the bottom we show the details of the loss function during optimization.}
        \label{overview}
\vspace{-0.4cm}
\end{figure*}

Unlike~\cite{vinker2023clipascene, frans2022clipdraw, schaldenbrand2022styleclipdraw}, optimization is performed for different regions in our method, which helps to highlight regions of interest, achieving coarse-to-concrete sketches. An intuitive and succinct learning process can be seen in Fig.~\ref{overview}. In each round of optimization, we are in pursuit of full content exploration rather than only the salient guidance. To achieve this, we present an edge-based stroke initialization and utilize a farthest point sampling (FPS) algorithm to uniformly sample strokes in the input image. Fig.~\ref{FPS-c} shows that our sampling method has a more efficient effect and is able to generate sketches in a reasonable manner. To transform a detailed scene into a simplified sketch, one must condense intricate visual elements into fundamental lines, shapes, and tones, all while retaining the scene's identifiable characteristics. We utilize intermediate layer of CLIP-ViT~\cite{dosovitskiy2020image} to guide the optimization, where encourages the creation of looser sketches that emphasize the scene’s semantics. We introduce a VGG16~\cite{simonyan2014very} model to promote the visual consistency and similarity between a sketch and an image.

We evaluate our proposed method for various photographs, including people, nature, indoor, animals et al., to showcase the effectiveness of our method. In summary, our contributions encompass four aspects: (1) We propose a novel method to convert a scene photograph into a sketch by optimizing it in a region-by-region fashion. A variety of sketching results can be achieved by adjusting regions in our scheme. (2) We apply a farthest point sampling (FPS) algorithm to the input image, evenly sampling positions on region edges as stroke positions, which are wisely utilized to emphasize content information. (3) We introduce two novel loss functions, a CLIP-based Semantic Loss and a VGG-based Feature Loss. These losses improve the generation of sketches, reflecting a balanced combination of both semantic and geometric features.

\section{Method}

We present a new method to processively convert a given scene photograph into a sketch with multiple rounds of optimization. An overview of our method can be seen in Fig.~\ref{overview}. Briefly, given an arbitrary image, our method can recursively learn its different regions by adding optimiziable Bézier curves. We define our sketch as some sets of black Bézier curves from different regions placed on a white background. Firstly, we introduce a stroke allocation method to reasonably divide the total number of strokes into different regions. Then we use a proposed stroke sampling to determine stroke locations, which will be converted into our initial strokes (Bézier curves). To improve convergence, we define the order from the global region to other regions, optimizing strokes successively.

\subsection{Stroke Initialization}
\textbf{Stroke Allocation.} Our method is based on multiple rounds of stroke superposition. Thus we first should consider the allocation of strokes in different regions.
As a case of fairness, we allocate strokes according to the edge points in each region by, based on the edge information gathered from contents of regions by edge detection~\cite{canny1986computational}. $Edge Detector$ in the following formula represents the edge extractor to gain the number of edge points in the region $R_i$. $r$ presents the number of regions. The stroke allocation ratio of region $R_i$ is calculated as follows:
\begin{equation}\label{E_i}
        {E_i}=Edge Detector(R_i)
\end{equation}
\begin{equation}\label{N_{E_i}}
        {N_{E_i}}=len(E_i)
\end{equation}
\begin{equation}\label{Ratio_i}
        {Ratio_i}=\frac{N_{E_i}}{\sum_i^r{N_{E_i}}}
\end{equation}

$N_s$ presents the number of total strokes. We then compute the final stroke allocation of region $i$:
\begin{equation}\label{N_i}
        {N_i}={Ratio_i}\cdot{N_s}
\end{equation}

\begin{figure*}[t]
    \centering
    \includegraphics[width=\textwidth]{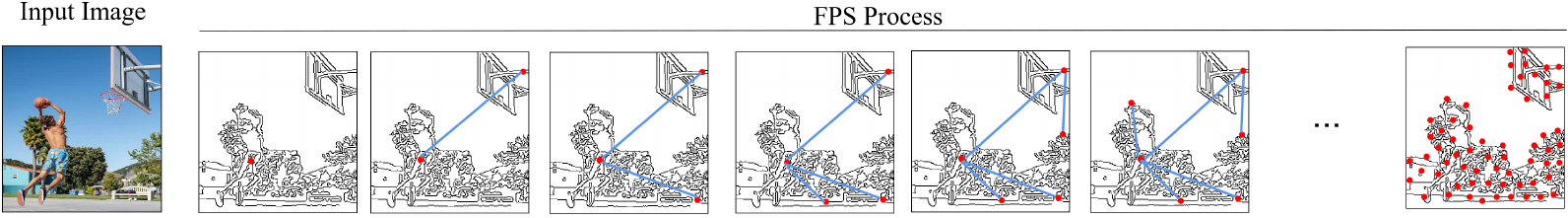} 
    \caption{An illustration of FPS process, which ensures that the sample points are well-spaced in the edge image.}
    \label{FPS}
    \vspace{-0.4cm}
\end{figure*}

\begin{figure*}[t]
    \centering
    \includegraphics[width=\textwidth]{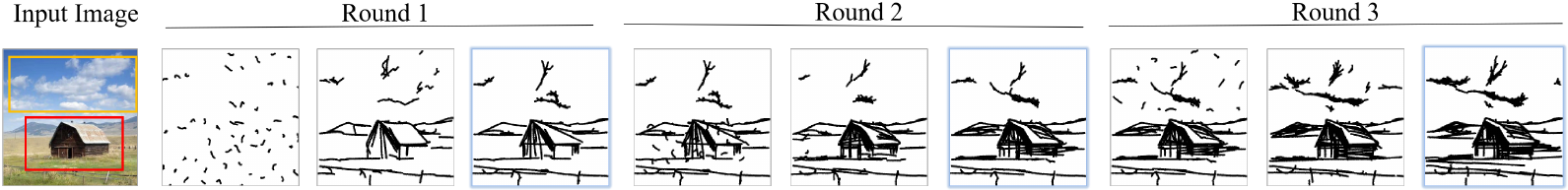} 
    \caption{The process of the optimization. The optimization results of the previous round will be superimposed with the initial strokes of the current region as the optimization input for the next round. }
    \label{optimization}
    \vspace{-0.4cm}
\end{figure*}

\begin{figure}[t]
    \centering
    \includegraphics[width=\columnwidth]{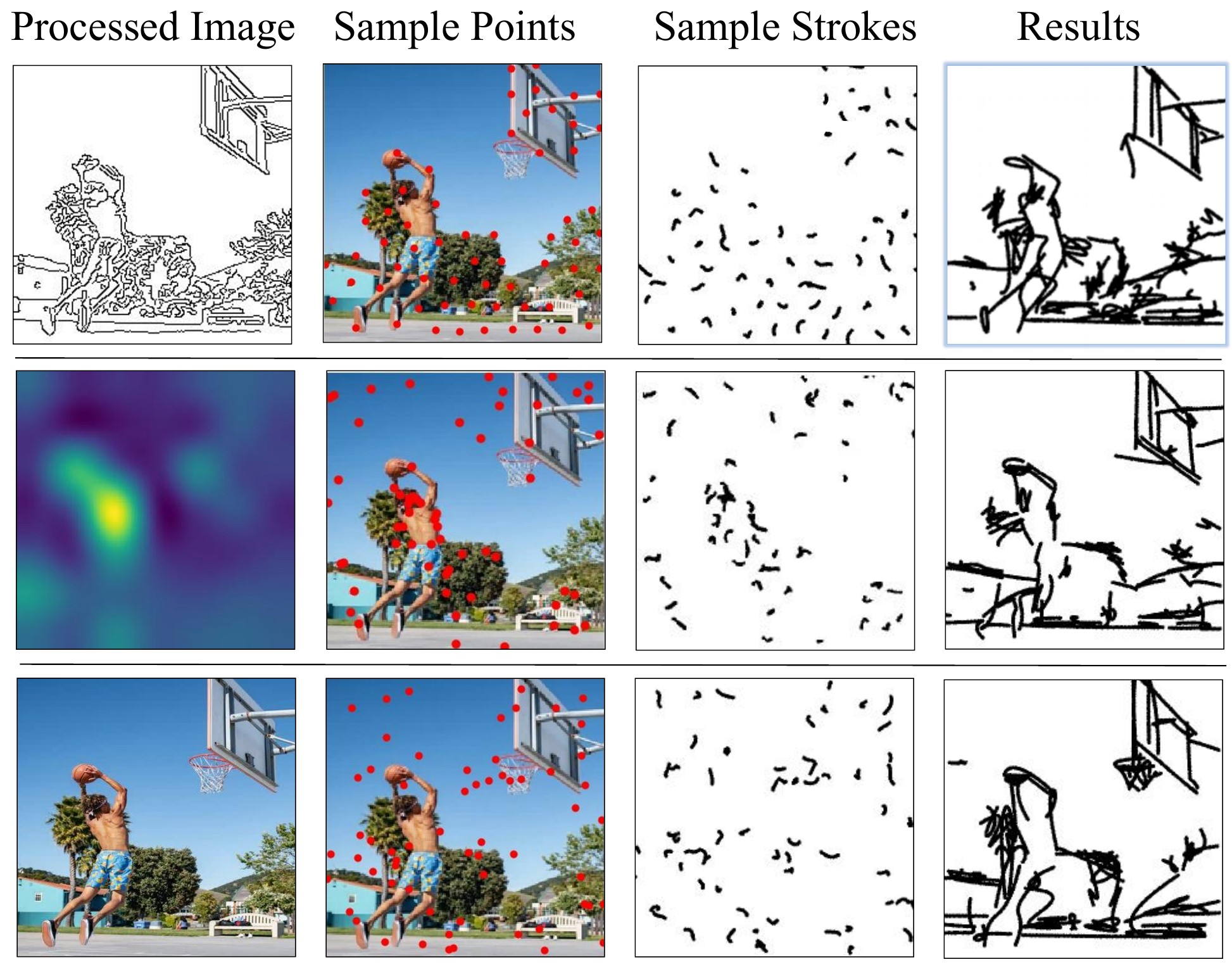} 
    \caption{Different Stroke Sampling Method. Top to Bottom: FPS sampling method, the sampling method of CLIPasso~\cite{vinker2022clipasso}, random sampling.}
    \label{FPS-c}
    \vspace{-0.4cm}
\end{figure}

\textbf{Stroke Sampling.} As each stroke corresponds to a sample point, the distribution of those points determines the distribution of strokes. For uniform coverage of image information and to reduce the lack of information, we adopt a farthest point sampling (FPS)~\cite{qi2017pointnet++} (the process shown in Fig.~\ref{FPS}). 
As a first step, we process the region to get the edges. Next, we use a farthest point sampling to sample strokes, selecting a subset of points from a larger set and maximizing the minimum distance between any two selected points. The process ensures that the selected points are well-spaced and representative of entire set. 

In Formula 5, we get a stroke sampling point set corresponding to the number of strokes in the region $R_i$.
\begin{equation}\label{ed}
         \{p_{1}, p_{2}, ..., p_{{n}}\}=EdgeDetector(R_i) 
\end{equation}
\begin{equation}\label{ni}
       \{p_{{k_1}}, p_{{k_2}}, ..., p_{{k_{N_i}}}\} = 
       FPS(\{p_{1}, p_{2}, ..., p_{{n}}\}, N_i)
\end{equation}

Fig.~\ref{FPS-c} shows that our stroke sampling method contributes significantly to the quality of the final sketch compared to the sampling method of CLIPasso~\cite{vinker2022clipasso} and random sampling. We further analyze the effect and variability of FPS in the supplementary material.

\subsection{Loss Function}

In previous works, some commonly used loss functions to minimize the error between images and results based on pixel-wise metrics. Although pixel-wise loss is simple yet intuitive, it is not sufficient to measure the distance between sketches and images as a sketch is highly sparse and abstract.
To address this, we leverage a pre-trained CLIP model to guide the training process.

\textbf{CLIP-based Semantic Loss.} For encoding shared information from both sketches and images, we follow CLIPasso~\cite{vinker2022clipasso} using CLIP model to compute the distance between the embeddings of the sketch $CLIP(Sketch)$ and image $CLIP(Image)$ as:
\begin{equation}\label{L_CLIP1}
        \mathcal{L}_{CLIP1}=dist(CLIP(Sketch), CLIP(Image))
\end{equation}

where $dist(x,y)=1-\frac{x\cdot y}{\|x\| \cdot\|y\|}$ is the cosine distance. And we are also based on the definition of the L2 distance loss function between certain activation layer $l$ in the CLIP model:
\begin{equation}\label{L_CLIP2}
    \mathcal{L}_{CLIP2} = {\left\|CLIP_{l}(Sketch)-CLIP_{l}(Image)\right\|_2^2}
\end{equation}
with $\lambda=0.1$, the CLIP-based Semantic Loss of the optimization is then defined as:
    \begin{equation}\label{L_CLIP}
         \mathcal{L}_{CLIP}= \mathcal{L}_{CLIP1}+\lambda \mathcal{L}_{CLIP2}
    \end{equation}

\textbf{VGG-based Feature Loss.} To further improve the structural similarity between the image and the sketch, we compute the L2 distance between the feature of them based on VGG16~\cite{simonyan2014very}:
\begin{equation}\label{Loss_s}
    \mathcal{L}_{VGG} = {\left\|VGG(Sketch)-VGG(Image)\right\|_2^2}
\end{equation}
More details about the CLIP model, activation layer and VGG model could be find in the supplementary material. The final objective of the optimization is then defined as:
    \begin{equation}\label{Loss_all}
         \mathcal{L}_{SUM}=\mathcal{L}_{CLIP} + \mathcal{L}_{VGG}
    \end{equation}

\begin{table*}[htbp]
\caption{Comparison of the LPIPS~\cite{zhang2018unreasonable}  score and SSIM~\cite{wang2004image} score. The score from left to right correspond to the results of our method, CLIPasso~\cite{vinker2022clipasso} and CLIPascene~\cite{vinker2023clipascene} based on sketch results at low and high abstraction levels.}
  \centering
    \begin{tabular}{c|c|c|c|c|c|c|c}
    
    \hline
          & Ours &CLIPasso & CLIPascene & XoG &  Photo-Sketching & Chan et al. & UPDG \\
    \hline
    LPIPS $\downarrow$ & 0.566 \quad 0.519 & 0.672 \quad 0.602 & 0.577 \quad 0.525 & 0.549& 0.728 & 0.520 & 0.640\\
    
    SSIM $\uparrow$ & 0.654  \quad 0.690 & 0.613 \quad 0.622 & 0.618 \quad  0.676 & 0.740 & 0.334 & 0.652 & 0.649\\
    \hline
    \end{tabular}
  \label{score}
  \vspace{-0.4cm}
\end{table*}

\begin{table*}[htbp]
\caption{Comparison of user preference rates. The scores from left to right correspond to the results of our method,
CLIPasso~\cite{vinker2022clipasso} and CLIPascene~\cite{vinker2023clipascene} are based on sketching results at low and high abstraction levels.}
  \centering
    \begin{tabular}{c|c|c|c|c|c|c}
    \hline
          & Ours v.s. CLIPasso & Ours v.s. CLIPascene & Ours v.s. XoG & Ours v.s. Photo-Sketching & Ours v.s. Chan et al. & Ours v.s. UPDG \\
    \hline
    Ours& 88.7\% \quad 96.3\% & 12.5\% \quad 38.6\% & 41.9\% & 96.2\%& 42.5\% & 40.0\% \\
   
    Others& 5.0\% \quad 2.5\% & 10.0\% \quad 26.4\% & 38.4\% & 3.8\% & 32.5\% & 28.8\% \\
  
    Equal &6.3\% \quad 1.2\% & 77.5\% \quad 60.0\% & 19.7\% & 0.0\% & 25.0\% & 32.2\% \\
    \hline
    \end{tabular}

  \label{userstudy}
  \vspace{-0.4cm}
\end{table*}

\subsection{Optimization}
In each optimization round, stroke parameters are overlaid with those from the previous round, with all parameters trained for 800 iterations. We start with a low-level sketch from the global region in the first round of optimization, refining it by incorporating strokes from other user-selected regions. As strokes from subsequent regions are sampled, each optimization round yields a flexible sketch. Fig.~\ref{optimization} illustrates this process, showcasing sketches at varying levels of abstraction across different optimization rounds. It should be noted that multiple rounds, while allowing for careful optimization of regions, can also lead to excessive overlap between regions, affecting the final result. 
\section{EXPERIMENTS AND RESULTS}

\subsection{Implementation}
\begin{figure*}[t]
        \centering
        \includegraphics[width=\textwidth]{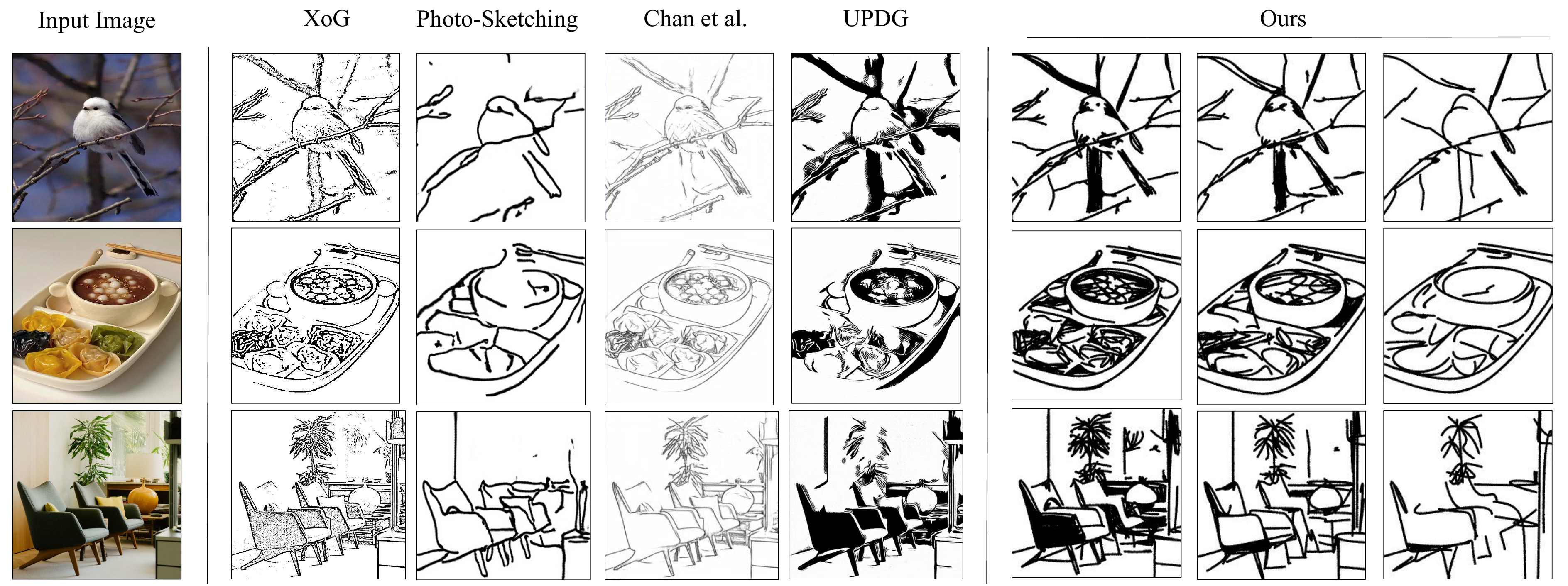} 
        \caption{Comparisons to methods that generate pixel-based sketches. On the right, are three representative sketches produced by our method depicting three levels of abstraction. More comparisons are provided in the supplementary material. More comparisons are provided in the supplementary material.}
        \label{compare}
\vspace{-0.4cm}
    \end{figure*}
\begin{figure*}[t]
        \centering
        \includegraphics[width=\textwidth]{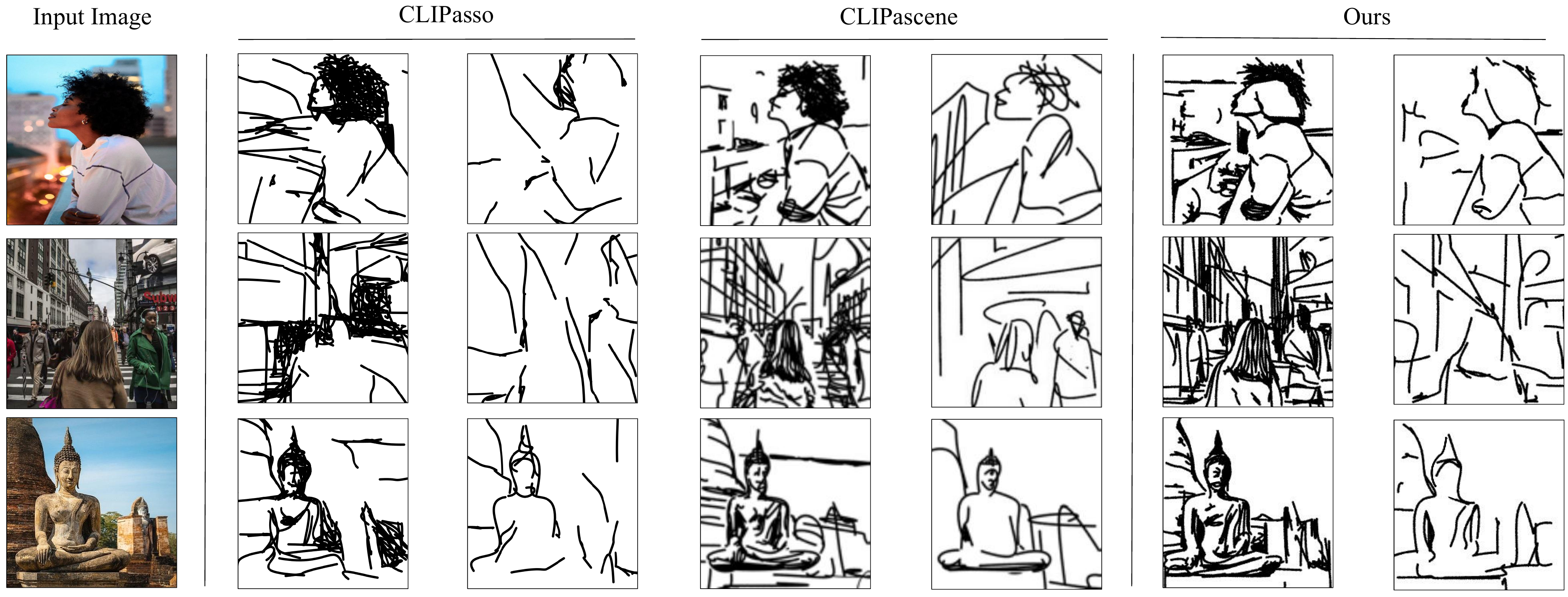} 
        \caption{Comparisons to methods that generate vector-based sketches. Representative sketches at high and low level of abstraction are generated.}
        \label{clip_compare}
\vspace{-0.4cm}
    \end{figure*}
\textbf{Optimization Details. }
We use Adam optimizer with a
learning rate set to 1. We evaluate the output sketch every iterations. Evaluation is done by computing the loss
without random augmentations. We repeat the optimization
process until convergence (when the difference in loss between two successive evaluations is less than 0.00001), this typically takes around 800 iterations. It takes 4 minutes to run 800 iterations on our server NVIDIA RTX 3090 GB, however, after 500 iterations, it is already possible to get a recognizable sketch for most images.

\textbf{Curves Details. }
For each curves, we sample positions for their first control points using the FPS algorithm and then randomly sample the next three control points of each Bezier curve within a small radius (0.05) of the first point.

\textbf{Edge Detection. }
We use the Canny edge detection~\cite{canny1986computational} on the preprocessed image. It takes the grayscale image as input along with two threshold values: the lower threshold and the upper threshold. We determining the appropriate threshold values as (20, 200) by a trial-and-error process.

\subsection{Evaluation Metrics} 
We calculate the Learned Perceptual Image Patch Similarity (LPIPS) ~\cite{zhang2018unreasonable} and Structural Similarity Index (SSIM) ~\cite{wang2004image} to evaluate models. The LPIPS assesses perceptual similarity by computing the L2 distance between deep feature representations extracted from pretrained convolutional neural networks, such as VGG~\cite{simonyan2014very}. It measures the perceptual quality and semantic similarity between the synthetic images and the ground-truth images. The SSIM evaluates structural similarity by analyzing luminance, contrast, and structure between image pairs, providing a value between 0 and 1, where 1 indicates identical images. 

\subsection{Comparison to State-of-the-art Methods} 

We conduct a comparison of our method against various alternative sketching techniques, such as XoG~\cite{winnemoller2012xdog}, Photo-Sketching~\cite{li2019photo}, Chan et al.~\cite{chan2022learning}, and UPDG~\cite{yi2020unpaired}. It is worth noting that none of these techniques possess the capability to adjust the abstraction level of the sketches or produce vector-based sketches. 
We also provide comparisons with vector-based methods CLIPasso~\cite{vinker2022clipasso} and CLIPascene~\cite{vinker2023clipascene}. 
Considering that the sketches generated by XoG~\cite{winnemoller2012xdog}, Photo-Sketching~\cite{li2019photo}, Chan et al.~\cite{chan2022learning} and UPDG~\cite{yi2020unpaired} are at different levels of abstraction, for a more comprehensive comparison, we use two different numbers of strokes (128 and 32) as inputs to our method, CLIPasso~\cite{vinker2022clipasso} and CLIPascene~\cite{vinker2023clipascene} to generate sketches at different levels of abstraction.
Quantitative results in Table~\ref{score} show our mehod achieves competitive LPIPS and SSIM scores compared to existing methods, indicating high perceptual similarity and structural consistency. While XoG achieves slightly better scores in SSIM (0.740) and LPIPS (0.549), its inability to control abstraction levels limits its flexibility compared to our method.

\begin{figure*}[t]
    \centering
    \includegraphics[width=\textwidth]{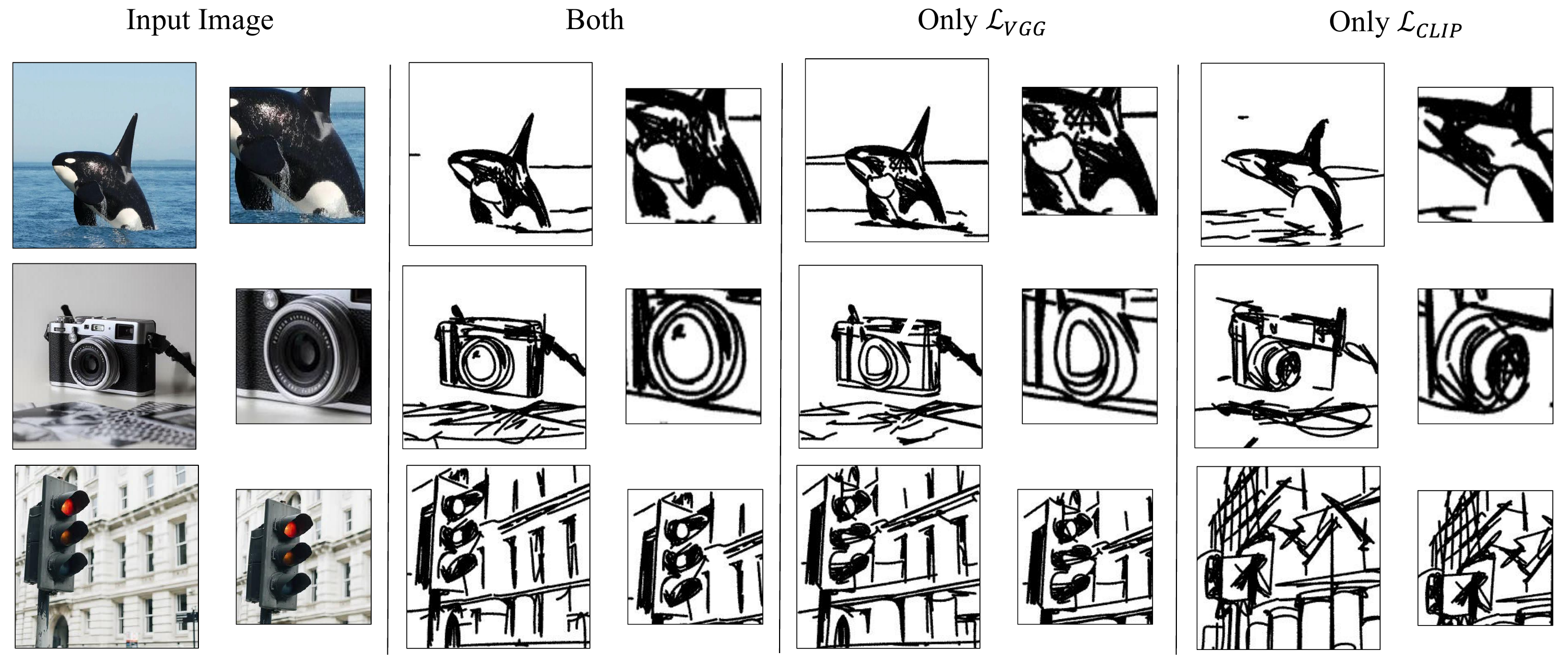} 
        \caption{Ablation of loss function. The effect of using only the CLIP-based Semantic Loss: capture abstract semantic details, like background, light and shadow. The effect of using only the VGG-based Feature Loss: emphasize structural features.}
        \label{ablation}
       
\end{figure*}
\subsection{User Study and Results}
\textbf{User study. }
To evaluate how well the sketches depict the input scene, we conducted a user study. We use 30 scene images to compare our sketches with six methods: CLIPasso~\cite{vinker2022clipasso}, CLIPascene~\cite{vinker2023clipascene}, XoG~\cite{winnemoller2012xdog}, Photo-Sketching~\cite{li2019photo}, Chan et al.~\cite{chan2022learning} and UPDG~\cite{yi2020unpaired}. 
The participants were presented with the input image along with two sketches, one produced by our method and the other by the alternative method. In order to make a fair study, we compared Photo-Sketching with our sketches at the lower abstraction level.  Table~\ref{userstudy} presents the final preference rates, showing MROSS gains higher acceptance from participants. 

\textbf{Results. }
Fig.~\ref{compare} and Fig.~\ref{clip_compare} demonstrate comparisons of MROSS with prior state-of-the-art methods. In Fig.~\ref{compare} we select XoG~\cite{winnemoller2012xdog}, Photo-Sketching~\cite{li2019photo}, Chan et al.~\cite{chan2022learning}, and UPDG~\cite{yi2020unpaired}, which do not have the ability to adjust the sketch abstraction level or generate vector-based sketches, for comparison. Given the different levels of abstraction of these methods, we choose to demonstrate the different abstraction level generation of our method by generating three different number of strokes (128, 64 and 32).
In Fig.~\ref{clip_compare}, we provide comparisons with vector-based methods CLIPasso~\cite{vinker2022clipasso} and CLIPascene~\cite{vinker2023clipascene}. Both can generate sketches at different abstraction levels and generate sketches in vector format. Therefore, we set the number of strokes at different abstraction levels (128 and 32) among these two methods for better comparison.

In contrast, MROSS, benefiting from alignment in structure and sematics, is able to maintain recognizability, underlying structure, and essential visual components of the scene drawn regardless of the level of abstraction. CLIPasso~\cite{vinker2022clipasso} aims to portray objects accurately. However, for the scene, it fails to depict the total image in details. 
Then we compared to CLIPascene~\cite{vinker2023clipascene}, for a fair comparison, we use the same decomposition technique to separate the input images into foreground and background. Then we use our method to sketch each part separately before combining them. CLIPascene~\cite{vinker2023clipascene} captures the whole image, which prevents them from emphasizing important regions of the input image. As shown in Fig.~\ref{clip_compare}, our method is able to capture the entire image while displaying contours and details of key objects.

\subsection{Ablation Study}
In Fig.~\ref{ablation} we present a comprehensive analysis of the results obtained by the results of various loss functions. Fig.~\ref{ablation} clearly shows that CLIP-based Semantic Loss can robustly capture semantic concepts but falls short in terms of structural fidelity. VGG-based Feature Loss could perfectly preserve the geometric structure, ensuring a faithful representation of the original form. Ablating these loss functions can prove that our final loss combination not only achieves semantic accuracy but also effectively preserves complex geometric details, achieving a harmonious balance between semantic expression and structural integrity.

\section{CONCLUSION}
We propose MROSS, an innovative scene sketching method that adopts a multi-round optimization strategy focusing on autonomously selected different regions. This enables a coarse-to-concrete representation, which is particularly beneficial for processing complex images. Meanwhile, we guide sketch generation with CLIP-based Semantic Loss and VGG-based Feature Loss. Extensive experiments and user studies confirm the superior scene generation performance of MROSS compared to existing methods.

\section{ACKNOWLEDGEMENT}
This research was supported by National Natural Science Foundation of China (Grant No., 62202152 \& 62202151).

\bibliographystyle{IEEEbib}
\bibliography{icme2025references}

\end{document}